\documentclass[aps,amsmath,pre,floatfix,superscriptaddress]{revtex4-2}
\usepackage{graphicx}
\usepackage{color}
\usepackage{geometry}
\usepackage{wrapfig}
\usepackage{comment}
\usepackage{amsmath}
\usepackage[export]{adjustbox}
\usepackage{float}
\usepackage{amssymb}
\usepackage[utf8]{inputenc}
\usepackage{afterpage}
\usepackage{lipsum}  
\usepackage{appendix}
\usepackage{natbib}
\usepackage{subcaption}
\usepackage{amsmath}
\usepackage{setspace}
\usepackage{tcolorbox}

\usepackage{hyperref}
\hypersetup{colorlinks=true, linktoc=all, linkcolor=blue, linktocpage, citecolor=blue}
\usepackage{caption}
\usepackage{geometry}\usepackage{geometry}

\usepackage[normalem]{ulem}
\graphicspath{{figs/}}

\usepackage[section]{placeins}

\begin{document}
\title{Do LLMs trust AI regulation? \\ Emerging behaviour of game-theoretic LLM agents}


\author{Alessio Buscemi$^{1,\dagger}$}
\author{Daniele Proverbio$^{2,\dagger}$}
\author{Paolo Bova$^{3,\dagger}$}
\author{Nataliya Balabanova$^{4}$}
\author{Adeela Bashir$^{3}$}
\author{Theodor Cimpeanu$^{5}$}
\author{Henrique Correia da Fonseca$^{6}$}
\author{Manh Hong Duong$^{4}$}
\author{Elias Fernández Domingos$^{7,8}$}
\author{António M. Fernandes$^{6}$}
\author{Marcus Krellner$^{5}$}
\author{Ndidi Bianca Ogbo$^{3}$}
\author{Simon T. Powers$^{9}$}
\author{Fernando P. Santos$^{10}$}
\author{Zia Ush Shamszaman$^{3}$}
\author{Zhao Song$^{3}$}
\author{Alessandro Di Stefano$^{3,\ddagger}$}
\author{The Anh Han$^{3,\ddagger,*}$}



 \maketitle
	{\footnotesize
		\noindent
        $^{1}$ Luxembourg Institute of Science and Technology\\
        $^{2}$ Department of Industrial Engineering, University of Trento\\
        $^{3}$ School Computing, Engineering and Digital Technologies, Teesside University\\
        $^{4}$ School of Mathematics, University of Birmingham\\
		$^{5}$  School of Mathematics and Statistics, University of St Andrews\\
        $^{6}$ INESC-ID and Instituto Superior Técnico, Universidade de Lisboa \\
        $^{7}$ Machine Learning Group, Universit\'e libre de Bruxelles\\ 
        $^{8}$ AI Lab, Vrije Universiteit Brussel\\
	    $^{9}$ Division of Computing Science and Mathematics, University of Stirling\\        
        $^{10}$ University of Amsterdam \\
        
\noindent $^\dagger$ Equally first authors \\
\noindent $^\ddagger$ Equally last authors \\
\noindent  $^\star$ Corresponding author: The Anh Han (T.Han@tees.ac.uk)
	}

  \maketitle

\section*{abstract}{ 
     There is general agreement that fostering trust and cooperation within the AI development ecosystem is essential to promote the adoption of trustworthy AI systems. By embedding Large Language Model (LLM) agents within an evolutionary game-theoretic framework, this paper investigates the complex interplay between AI developers, regulators and users, modelling  their strategic choices under different regulatory scenarios. Evolutionary game theory (EGT) is used to quantitatively model the dilemmas faced by each actor, and LLMs provide additional degrees of complexity and nuances and enable repeated games and incorporation of personality traits. Our research identifies emerging behaviours of strategic AI agents, which tend to adopt more "pessimistic" (not trusting and defective) stances than pure game-theoretic agents. We observe that, in case of full trust by users, incentives are effective to promote effective regulation; however, conditional trust may deteriorate the "social pact". Establishing a virtuous feedback between users' trust and regulators' reputation thus appears to be key to nudge developers towards creating safe AI. However, the level at which this trust emerges may depend on the specific LLM used for testing. Our results thus provide guidance for AI regulation systems, and help predict the outcome of strategic LLM agents, should they be used to aid regulation itself. 
    }

\vspace{3mm}

\noindent\textbf{Keywords:} AI governance, AI regulation, trustworthy AI, game theory, LLM, behavioural dynamics.

\section{Introduction}

As Artificial Intelligence (AI) applications become widespread, debates are taking place about how to regulate it \cite{ bengio2024managing,floridi2018ai4people, baker2023executive, finocchiaro2024regulation,hammond2025multiagentrisksadvancedai,bengio2025international}. In the ideal scenario, AI systems should be trustworthy and safe, so that users can trust them and enable broad and safe adoption. Regulation is typically viewed as a key way to achieve such desired outcomes \cite{powers2024stuff}, and regulators are developing acts and guidelines, such as the EU AI Act, to balance trustworthiness, safety and access to innovation \cite{laux2024trustworthy}. Nonetheless, key questions remain as to which levels of restrictions should be placed upon AI developers, about who should create and enforce such regulations, who may be better suited to perform monitoring, and in general which actions could better drive trust-building among users \cite{siegmann2022brussels, tallberg2023global, clark2019regulatory, anderljung2023frontier, clarke2019regulatory}. 

Albeit the discourse around the topic is mostly qualitative and limited in its formulation of formal predictions \cite{dafoe2023ai, hadfield2023regulatory, zaidan2024ai}, recent attempts have been developed to create quantitative and systematic frameworks to analyse the mutual relationships among all involved actors, and to predict the effect of different regulatory systems using game-theoretic principles \cite{alalawi2024trust, kondor2024complex, chan2024balancing,bova2023both,han2020regulate}. The typical framework considers asymmetric games among three actors -- regulators, AI developers and users -- with different decision-making strategies and dilemmas. In fact, each actor has risk-minimising (or profit-maximising) goals: users can decide whether to trust and adopt AI systems \cite{eurobarometer, ipsos}, depending on whether they consider them trustworthy or not, and therefore run the risk that such systems do not ensure their users' interests or may be even malicious \cite{han2021or, buscemi2024roguegpt,andras2018trusting}. This, in turn, derives from AI developers typically working in competition with each other and pursuing their own interest over compliance to regulation and trust-building towards users \cite{armstrong2016racing, askell2019role,cottier2024who}. Regulators also need to balance the protection of users' rights and the management of resources, \textit{e.g.}, by delegating the monitoring to private audit companies \cite{cihon2021ai, north1990institutions}. Using evolutionary game theory enables us to predict  long-term behavioural outcome of interactions among the three actors and to estimate the effect of incentives for regulators.

On top of employing well-established methods from evolutionary game-theory, which embed human decision-making processes in a formal and proven domain \cite{hofbauer1998evolutionary, sigmund2010calculus}, we may also exploit the newest technologies to obtain complementary models and predictions. Indeed, AI systems themselves have been suggested to enable suitable replicas of human actions \cite{park2023generative, bail2024can}: hence they could, in principle, be used to perform strategic games involving complex, non-linear and multi-faceted agents. We thus create a new framework to experiment the regulatory dynamics among AI agents, behaving as three actors (regulators, AI developers and users, as in \cite{alalawi2024trust}) in a regulatory ecosystem. We blend AI agents within a game-theoretic setting: three AI agents interact dynamically, after being prompted in such a way to represent the three desired actors. To this end, we use the new generation of Large Language Models (LLMs) \cite{zhao2023survey}. As it was observed that different LLMs may produce contrasting results in various tasks \cite{buscemi2024large, buscemi2024chatgpt, lee2024evaluating}, we employ two different models: GPT-4o from OpenAI's GPT family \cite{chatgpt} and Mistral Large by Mistral \cite{mistral}. The dilemmas that each actor typically faces are then embedded in the form of payoff matrices, in the spirit of evolutionary game theory. This enables direct comparison with previous results and ensure reproducibility  and interpretability of the results. We also consider the possibility \cite{north1990institutions} that users may condition their trust on the effectiveness of regulators, thereby representing additional incentives for regulators to comply with users' needs. Moreover, since using LLMs enables greater flexibility  for capturing inherent characteristics of individuals \cite{lu2024llms,SONG2025245}---which experimental evidence suggests to significantly impact human behaviours   \cite{van2013psychology}---we  test whether suggesting specific personalities to AI agents modifies significantly the emerging dynamics.

Our experiments have three main purposes. First, to observe the emerging behaviours of strategic AI agents, which are requested to model social interactions in a complex evolutionary game setting. Second, to compare such behaviours with those expected from game-theoretic predictions, so as to validate and interpret them under the lenses of a known theory, thereby improving interpretability of the emerging outcomes. Third, we address the recent deployment of AI agents at all organizational levels: in the scenario where expert AI agents are employed by private organizations and developers to help draft governance guidelines, how may they respond? Our study provides first systematic predictions to such questions, by considering one-shot and repeated games, and by explicitly including personality traits in the AI agents.

We observe that LLMs provide results that are more nuanced and not always aligned to game-theoretic predictions. This may be associated to LLM having additional complexity coming from the training process. Overall, we observe a variety of strategies emerging from combinations of payoffs and scenarios; in general, having conditional trust seems to promote defective stances by developers and regulators, while AI-users tend not to eventually trust them. Instead, full trust by AI-users promote higher chances of virtuous behaviours by the other agents. In general, GPT-4o has a more optimistic attitude than Mistral Large, which highlights potential inconsistencies and sensitivity to payoffs among LLMs -- which, in turn, should promote studies to improve the reproducibility of tests.

In the next section, we introduce the game-theoretic payoff settings, the setup of the LLM agents and the characteristics of the tests. Results and discussion for each analysis and research purposes will follow.

\section{Methods}
\label{sec:methods}

\subsection{The three-actors game-theoretic setting}
\label{sec:game_setting}

\begin{figure}
\includegraphics[width=0.7\linewidth]{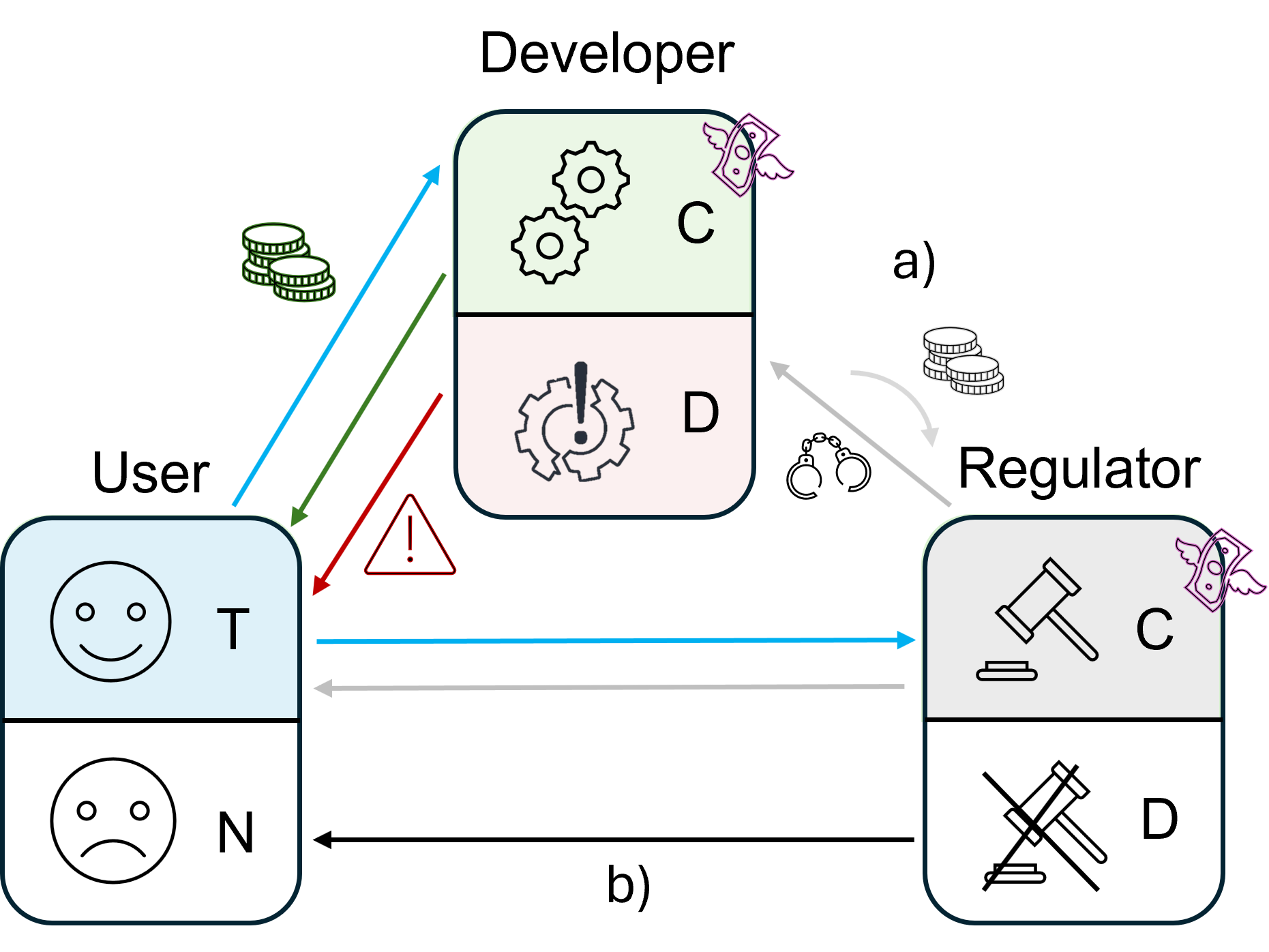}
\caption{Scheme of the core features for the three-player interaction model. Users may trust or not; if they do ($T$), other agents get benefits (blue lines). If they do not ($N$), no adoption is enacted and other agents get no benefits. Developers may comply ($C$) with regulations and develop safe AI which, if adopted, yields benefits for users (green line); however, it may be costly. Instead, developers may create unsafe AI ($D$) which, if adopted, may yield partial or negative payoff to users (red line). Regulators may be strict ($C$), using resources but gaining benefits if they catch defective developers (a), or lenient ($D$). If users have access to regulators' reputation (b), they can decide whether to trust conditionally.}
\label{fig:scheme}
\end{figure}

We build our analysis upon the model from \cite{alalawi2024trust}, which formalises the multi-party interactions among three actors in a regulatory ecosystem in the form of a game (see Fig. \ref{fig:scheme}). The interactions model the simplest form of regulatory ecosystem identified by \cite{powers2024stuff}: each actor represents the average behaviour of a population of AI users, developers and regulators, which mutually influence one another in a direct and non-mediated fashion. This model is rather simplistic and does not include influences from, \textit{e.g.}, culture or economy, nor the mediating effect of other actors such as scholars or the media \cite{powers2024stuff,balabanova2025media}; nonetheless, it includes the key elements considered during the development processes, and is similar to typical models of regulations for information systems such as the General Data Protection Regulations (GDPR) and the AI Act of the European Union. Each actor can chose among binary options in an  interaction with other actors: Users can trust ($T$) or not ($N$) an AI system; developers either comply ($C$) or not ($D$) with regulations; regulators enforce compliance ($C$) or not ($D$). 

Each encounter or interaction is framed as a game, whose outcome depends on the strategy of each actor. The strategy is dictated by the weights placed upon each action, summarised in the game payoff matrix. The payoff matrix is built as described in \cite{alalawi2024trust}. The parameters associated with each strategy payoff for each user are described as follows. 

\begin{itemize}
    \item \textbf{Users:} users trust (T) or not (N) an AI system, depending on a combination of their trust in regulators and in developers. If $T$ when both regulators and developers cooperate, users get a benefit $b_U$; if developers defect (do not comply), users use unsafe AI and gain reduced or even negative benefits, that is, $\epsilon \times b_u$, with $\epsilon \in (-\infty; 1]$ represents a risk factor. 
    \item \textbf{Creators:} developers may comply with regulations ($C$) or defect ($D$). They gain a benefit $b_P$ from selling the product (i.e. if users trust and adopt AI), and incur an additional cost $c_P$ to create safer AI (while the cost of creating unsafe AI is normalised to $0$). However, if the regulator enforces rules and developers play $D$, they suffer institutional punishment resulting in a payoff loss  of $u$.
    \item \textbf{Regulators:} they can either enforce compliance ($C$) o be lenient ($D$). We assume that regulators earn a benefit $b_R$ when the user trusts and adopts AI, \textit{e.g.} by being funded by taxes on the sales of AI products, or by increasing investments in regulation by governments when there is higher uptake of AI. On the other hand, creating rules and monitoring technologies is costly; playing $C$ thus requires an extra cost $c_R$, while $D$ has a cost normalised to 0. Finally, administering the punishment is associated with an additional cost $v$, but cooperative regulators are rewarded with a benefit $b_{fo}$ by governments when finding out a defective developer.
\end{itemize}

We consider two cases for users to place their trust upon regulators: if regulators’ reputation is publicly available, users can adjust their trust depending on whether the regulators’ reputation is good or not. This leads to a Conditional Trust (\textit{CT}) scenario. Otherwise, this reputational  information may not be available, and thus users place their trust solely depending on their perceived benefits. For each scenarios, we test the influence of $b_{fo}$ by sampling a subset of the values from \cite{alalawi2024trust}, to enable direct comparison while employing reasonable resources (cost and time) when performing the experiments with LLMs.

The payoff matrices, with and without \textit{CT}, reproduce those in \cite{alalawi2024trust} and are given in Tables \ref{tab:CT_payoff} and \ref{tab:nonCT_payoff}, respectively.

\begin{table}[h!]
    \centering
    \normalsize
    \caption{AI Governance model with  with incentives for regulators and conditional trust. Agents are: User \textit{Us.}, Creator \textit{Cr} and Regulator \textit{Re}. Users act conditionally (\textit{CT}) on the regulators' reputation, assumed to be publicly available before the game. }
    \label{tab:CT_payoff}
        \begin{tabular}
        {|p{0.045\textwidth}|p{0.045\textwidth}|p{0.045\textwidth}|| p{0.2\textwidth} | p{0.2\textwidth}| p{0.2\textwidth}|}
       \hline
       \multicolumn{3}{|c||}{\textbf{Strategies}} &  \multicolumn{3}{ c |}{\textbf{Payoffs}} \\
          \hline
        \textbf{Us.} & \textbf{Cr} & \textbf{Re} & \textbf{User} & \textbf{Creator} & \textbf{Regulator} \\
        \hline \hline
         \textit{CT} & \textit{C} & \textit{C} & $b_U$ & $b_P-c_P$ & $b_R- c_R$ \\
        \hline
         \textit{CT} & \textit{C} & \textit{D} &  {$0$} &  {$- c_P$} &  {$0$}\\
        \hline
         \textit{CT} & \textit{D} & \textit{C} & $\varepsilon b_U$ & $b_P-u$ & $b_R-c_R-v+b_{fo}$\\
        \hline
         \textit{CT} & \textit{D} & \textit{D} &  {$0$} &  {$0$} &  {$0$} \\
        \hline
         \textit{N} & \textit{C} &\textit{C} & 0 & $-c_P$ & $-c_R$\\
       \hline
         \textit{N} & \textit{C} & \textit{D} & 0 & $- c_P$ & 0 \\
       \hline
        \textit{N} & \textit{D} &\textit{C} & 0 & 0 & -$c_R$\\
       \hline
         \textit{N} & \textit{D} & \textit{D} & 0 &  0 &  0\\
      \hline
     \end{tabular}
\end{table}

\begin{table}[h!]
    \centering
    \normalsize
    \caption{AI Governance model with incentives for regulators, but without conditional trust. Agents are: User \textit{Us.}, Creator \textit{Cr} and Regulator \textit{Re}). Users trust ($T$) solely based on their payoff and \textit{a-priori} attitude.}
    \label{tab:nonCT_payoff}
        \begin{tabular}
        {|p{0.045\textwidth}|p{0.05\textwidth}|p{0.05\textwidth}|| p{0.2\textwidth} | p{0.2\textwidth}| p{0.2\textwidth}|}
       \hline
       \multicolumn{3}{|c||}{\textbf{Strategies}} &  \multicolumn{3}{ c |}{\textbf{Payoffs}} \\
          \hline
        \textbf{Us.} & \textbf{Cr} & \textbf{Re} & \textbf{User} & \textbf{Creator} & \textbf{Regulator} \\
        \hline \hline
         \textit{T} & \textit{C} & \textit{C} & $b_U$ & $b_P-c_P$ & $b_R- c_R$ \\
        \hline
         \textit{T} & \textit{C} & \textit{D} & $b_U$ & $b_P - c_P$ & $b_R$\\
        \hline
         \textit{T} & \textit{D} & \textit{C} & $\varepsilon b_U$ & $b_P-u$ & $b_R-c_R-v {+b_{fo}}$\\
        \hline
         \textit{T} & \textit{D} & \textit{D} & $\varepsilon b_U$ & $b_P$ & $b_R$ \\
        \hline
         \textit{N} & \textit{C} &\textit{C} & 0 & $-c_P$ & $-c_R$\\
       \hline
         \textit{N} & \textit{C} & \textit{D} & 0 & $- c_P$ & 0 \\
       \hline
        \textit{N} & \textit{D} &\textit{C} & 0 & 0 & -$c_R$\\
       \hline
         \textit{N} & \textit{D} & \textit{D} & 0 &  0 &  0\\
      \hline
     \end{tabular}
\end{table}
Finally, we consider two settings for the games: a series of one-shot games, where each encounter happens once and delivers a result, and a repeated games, where players interact more than once and may adjust their behaviour depending on past direct interactions.
It has been established in other repeated game settings such as the Prisoner's Dilemma, the Public Good Games, and the AI race interactions, that when the interaction among the same pair or group of players are repeated, desirable behaviours such as cooperation and safe development become more frequent via direct reciprocity \cite{nowak2006five,van2012emergence,han2020regulate}.
We test if repeated interactions in this three-party game will improve desirable outcomes.

\subsection{AI agents setup}
\label{sec:AI_ag_setup}

The games are set using LLM agents whose payoffs are given as described above. To setup agents within a game-theoretic framework, we employ the Framework for AI Agents Bias Recognition using Game Theory (FAIRGAME) \cite{buscemi2025fairgame}. 
FAIRGAME enables testing of user-defined games, described in textual format and incorporating any desired payoff matrix.
Additionally, it allows for the specification of agent traits that will participate in these games.
The agents can be instantiated using any LLM of choice by invoking the corresponding APIs.

To run, FAIRGAME requires the following inputs:
\begin{itemize}
    \item \textbf{Configuration File:} A file that defines the setup of both the agents and the game. The default format is JSON. In this study, we use custom files listing the parameters and payoff weights associated with the game, as described above.
    \item \textbf{Prompt Template:} A text file that defines the instruction template, providing a literal description of the game. It includes placeholders that are dynamically populated with information from the configuration file at each round, ensuring customization for each agent. The template used for all experiments is available in Supplementary Section S1.
\end{itemize}

\begin{table}[h!]
    \centering
    \normalsize
    \caption{Parameters provided to FAIRGAME.}
    \label{tab:fairgame}
    \begin{tabular}{|p{0.5\textwidth}|p{0.3\textwidth}|}
        \hline
        \textbf{Parameter} &  \textbf{Value} \\
        \hline
            Number of agents & 3 \\
        \hline
            Names of the agents & regulator; developer; user \\
        \hline
            Personalities of the agents & None; None; None \\
         \hline
            Underlying LLM & OpenAI GPT-4o; Mistral Large \\
        \hline
            Number of rounds & 1 (for one-shot games); 10 (for repeated games) \\
        \hline
            Agents communicate & False \\
         \hline
            Agents know the personalities of the others & False \\    
        \hline
            Stopping condition & None \\    
        \hline
        \end{tabular}
\end{table}

Table~\ref{tab:fairgame} lists all the parameters used to run the main experiments (see Sections  \ref{sub:one_shot_results}, \ref{sub:repeated_results}), as defined in the configuration file.
The framework employs three distinct agents, each assigned a specific role: regulator, developer, and user. 

The underlying language model for the agents is either OpenAI's GPT-4o or Mistral Large. Once a model is selected, it remains consistent across all agents in a given simulation. That is, we conduct simulations where all agents are based on GPT-4o and others where all agents are based on Mistral Large, but we do not mix models within the same game.

We conduct both one-shot games, consisting of a single round, and repeated games, which span ten rounds. Notably, the agents operate independently, with no direct communication among them, ensuring that all decisions are made in isolation. Moreover, the agents lack prior knowledge of the personalities or strategic tendencies of the other participants. The games proceed without predefined stopping conditions and continues for the designated number of rounds.

Although the framework supports the specification of agent personalities, in the main experiments, all agents operated without assigned personalities, maintaining their default behaviour.
This guarantees that decisions are made solely according to their respective roles, adhering to the default behaviour of the LLMs without introducing additional factors. We also construct additional experiments to specifically inquire the effect of setting agents' personalities: in Section~\ref{sub:personality_results}, we conduct a subset of the main experiments to examine how contrasting personalities influence the outcomes.
Specifically, we test the following personality traits, fully described to LLMs through the prompt template file to avoid ambiguity. 
\begin{itemize}
    \item \textbf{User}: risk-adverse, i.e. you reject new AI systems to avoid uncertainty OR risk-taking, i.e. you adopt new AI systems to benefit from potential advancements.
    \item \textbf{Developer}: aggressive, i.e. you develop quickly to stay ahead, accepting some risks OR cooperative, i.e. you take a cautious approach to minimize risk.
    \item \textbf{Regulator}: lenient, i.e. you trust developers to regulate themselves OR strict, i.e. you require verification before deployment to ensure safety.
\end{itemize}

\section{Results}

\subsection{One-shot games with and without conditional trust}
\label{sub:one_shot_results}
We first consider the case of a set of one-shot games, and the effect of incentives for regulators ($b_{fo}$) on the adoption of specific strategies. 
We will compare the results with outcomes from evolutionary game dynamics in finite populations, which is known to elicit notable stochastic effects on evolutionary outcomes due to, \textit{e.g.}, errors in social learning and behavioural exploration \cite{nowak2004emergence}.
This stochastic approach has proven to be effective in explaining human behaviours in controlled experiments \cite{zisis2015generosity,rand2013evolution}.

Using AI agents already embeds stochasticity, due to the very nature of LLMs which always contains a degree of randomness \cite{vidler2025playing} and a condensation of stochastic outcomes into the output Softmax functions \cite{deng2024zero}. In turn, this allows us to directly compare the results with game-theoretic stochastic approaches, which are known to help in explaining empirical observations from human behavioural experiments and have been thoroughly investigated in the literature \cite{rand2013evolution,zisis2015generosity}. As a consequence, we can observe the emergence of preferred strategies in games with three AI agents, and directly compare them with the predictions from evolutionary game theory.

\begin{figure}[h]
\includegraphics[width=\linewidth]{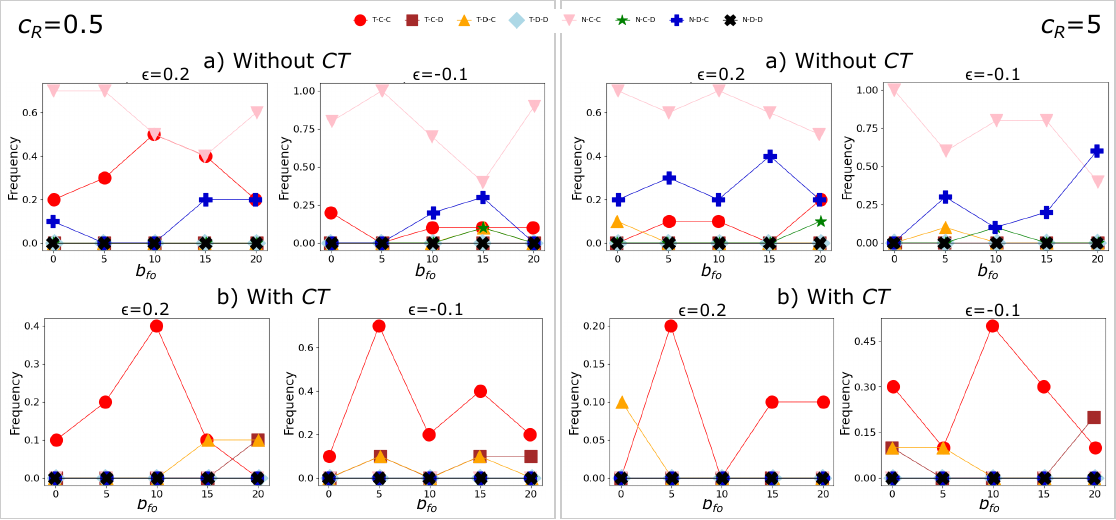}
\caption{Results for the one-shot game, using GPT-4o. Left box: low regulation cost ($c_R$ = 0.5). Right box: high regulation cost ($c_R = 5$). Each panel corresponds to a different value for $\epsilon$, \textit{i.e.}, the risk for users to adopt unsafe AI ($\epsilon<0$ has higher risk). Conditional trust promotes full trust, cooperative regulation and safe development. Parameters set to: $b_U$ = $b_R$ = $b_P$ = 4, $u$ = 1.5, $v$ = 0.5, $c_P$ = 0.5.}
\label{fig:one_shot_GPT}
\end{figure}

\begin{figure}[h]
\includegraphics[width=\linewidth]{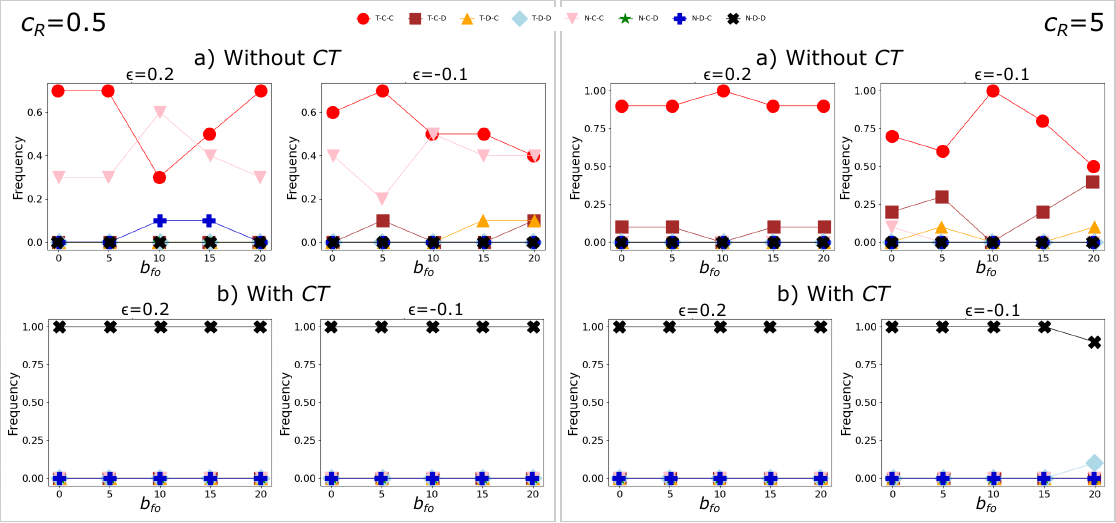}
\caption{Results for the one-shot game, using Mistral Large. Left box: low regulation cost ($c_R$ = 0.5). Right box: high regulation cost ($c_R = 5$). Each panel corresponds to a different value for $\epsilon$, \textit{i.e.}, the risk for users to adopt unsafe AI ($\epsilon<0$ has higher risk). Conditional trust yields low trust. Parameters set to: $b_U$ = $b_R$ = $b_P$ = 4, $u$ = 1.5, $v$ = 0.5, $c_P$ = 0.5.}
\label{fig:one_shot_Mistral}
\end{figure}

Figs. \ref{fig:one_shot_GPT} and \ref{fig:one_shot_Mistral} summarise the results, for games testing different values of risk scores $\epsilon$ and regulation cost $c_R$, with and without conditional trust ($CT$). Fig. \ref{fig:one_shot_GPT} refers to tests conducted with GPT-4o-based LLM agents, while Fig. \ref{fig:one_shot_Mistral} uses Mistral. We show the frequencies of the eight ($2^3$) possible combinations of strategies for the three players, depending on the reward for cooperative regulators $b_{fo}$. We consider $\epsilon = -0.1$ when defection by the developers results in highly detrimental outcome for the users, and $\epsilon  = 0.2$ when a lower, but still positive value to the adoption of unsafe AI  is assumed (for instance, when it anyway allows  access to some services).

When trust is not conditioned on the regulators' reputation, GPT-based agents tend to prefer a situation where users do not trust AI (except for the case where regulators have a low regulation cost and users still gain some benefit by adopting unsafe AI, upper-left panel of Fig. \ref{fig:one_shot_GPT}, where the strategy with full trust and compliance coexists for medium incentives for regulators). Regulators have here the tendency to comply more frequently, especially with a higher benefit for catching unsafe developers, $b_{fo}$. On the other hand, we observe a mixing of cooperation and defection by developers. Moreover, we typically observe that, if regulators have incentives to catch defective developers, their complying proportion increases when the proportion of defective developers increases, even at higher costs of regulation. This effect thus have the capability of potentially coping with developers' behaviours. These findings are quite different from what happens in a purely game-theoretic setting (compare with Figs. 6 and 7 of \cite{alalawi2024trust}), where the TDD strategy was usually dominant. Interestingly, this may be explained by the fact that GPT-4o was trained on recent real-world data, which show an overall tendency of users to trust AI only partially \cite{eurobarometer,ipsos}, of regulators to try and navigate the regulatory landscape with new actions (think \textit{e.g.} of the EU AI Act), and of developers to display a variety of approaches, ranging from the safety guardrails of OpenAI and Microsoft to the more lax attitude of developers such as xAI with its Grok 3. It thus seems that GPT-based agents mix the pure payoff-based results with statistical outcomes derived from empirical data.

Instead, with Conditional Trust, GPT-based agents tend to be more trusting and complying, both for low and high regulation cost. In the first case, this is in line with game-theoretic results, while in the second case it shows the overall "positivity" of the LLM towards users', developers' (and thus regulators') behaviour, which tend to become defective only with high cost despite the incentives. Even in this case, we can recognise some effect of the training data on the game's output, which thus mixes statistical evidence about the real world and payoff-based strategies.

On the other hand, Mistral behaves rather differently from GPT. The scenario without \textit{CT} and with $c_R=0.5$ is very close to GPT's one, but then the LLM diverges in its outcome. Differently from GPT, it remains "optimistic" (preferring the TCC strategy) also when the regulators' cost is higher, in case of full trust. Instead, conditional trust triggers a NDD scenario; in the case of high $c_R$, it is consistent with observations made using pure game theory (see Fig. 7 in \cite{alalawi2024trust}), while the same occurrence for lower $c_R$ suggests that Mistral views conditional trust as an overall detrimental element. The fact that \textit{CT} has such a prominent effect on Mistral's outcomes suggests that the model is very sensitive to the changes made in the payoff matrix to accommodate the \textit{CT} mechanism.

\subsection{Repeated games}
\label{sub:repeated_results}

\begin{figure}[h]
\includegraphics[width=\linewidth]{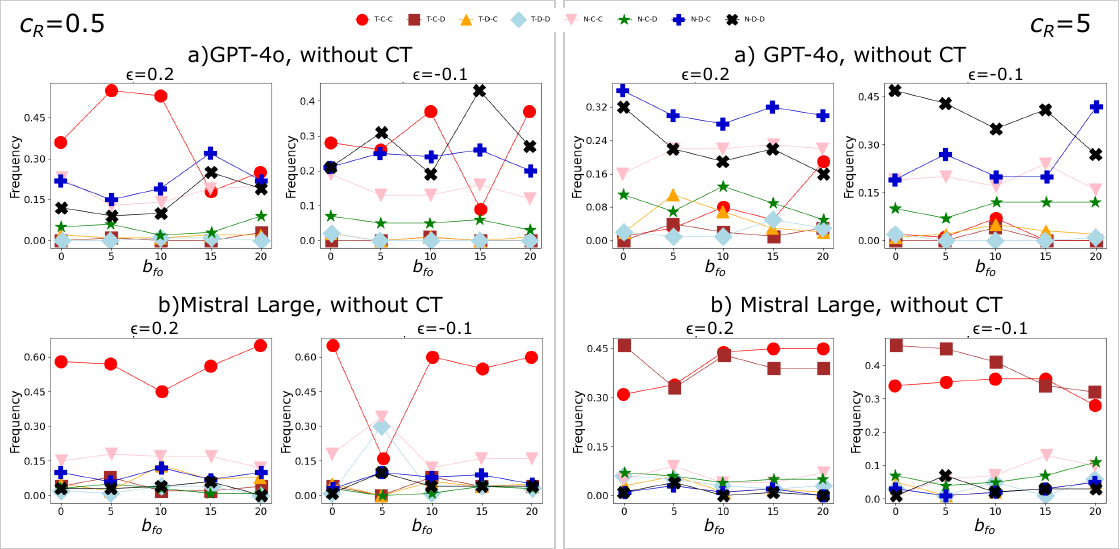}
\caption{Results for the repeated games over 10 rounds (average over rounds), using GPT-4o and Mistral Large. Left box: low regulation cost ($c_R$ = 0.5). Right box: high regulation cost ($c_R = 5$). Each panel corresponds to a different value for $\epsilon$, \textit{i.e.}, the risk for users to adopt unsafe AI ($\epsilon<0$ has higher risk). }
\label{fig:repeated_both}
\end{figure}

\begin{figure}[h]
\includegraphics[width=\linewidth]{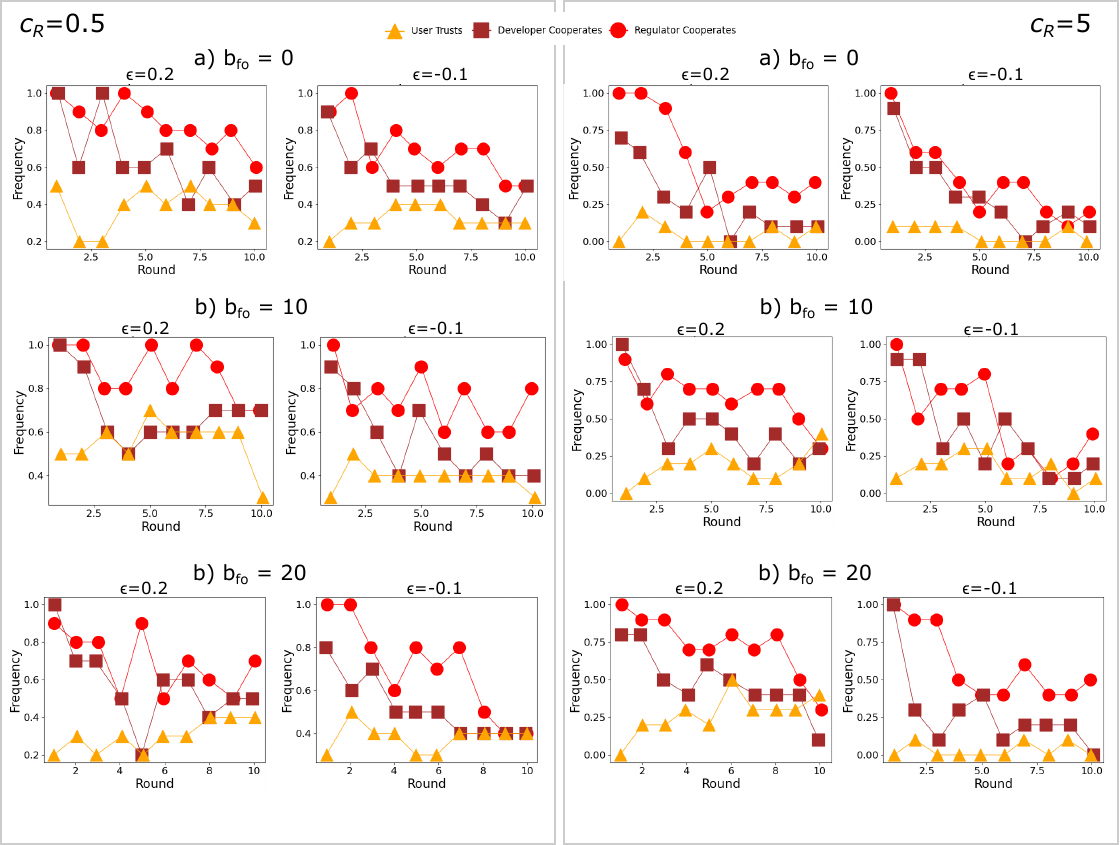}
\caption{Results for the repeated games over each of the 10 rounds, for different values of $b_{fo}$, using GPT-4o. Left box: low regulation cost ($c_R$ = 0.5). Right box: high regulation cost ($c_R = 5$). Each panel corresponds to a different value for $\epsilon$, \textit{i.e.}, the risk for users to adopt unsafe AI ($\epsilon<0$ has higher risk). }
\label{fig:repeated_GPT_rounds}
\end{figure}

\begin{figure}[h]
\includegraphics[width=\linewidth]{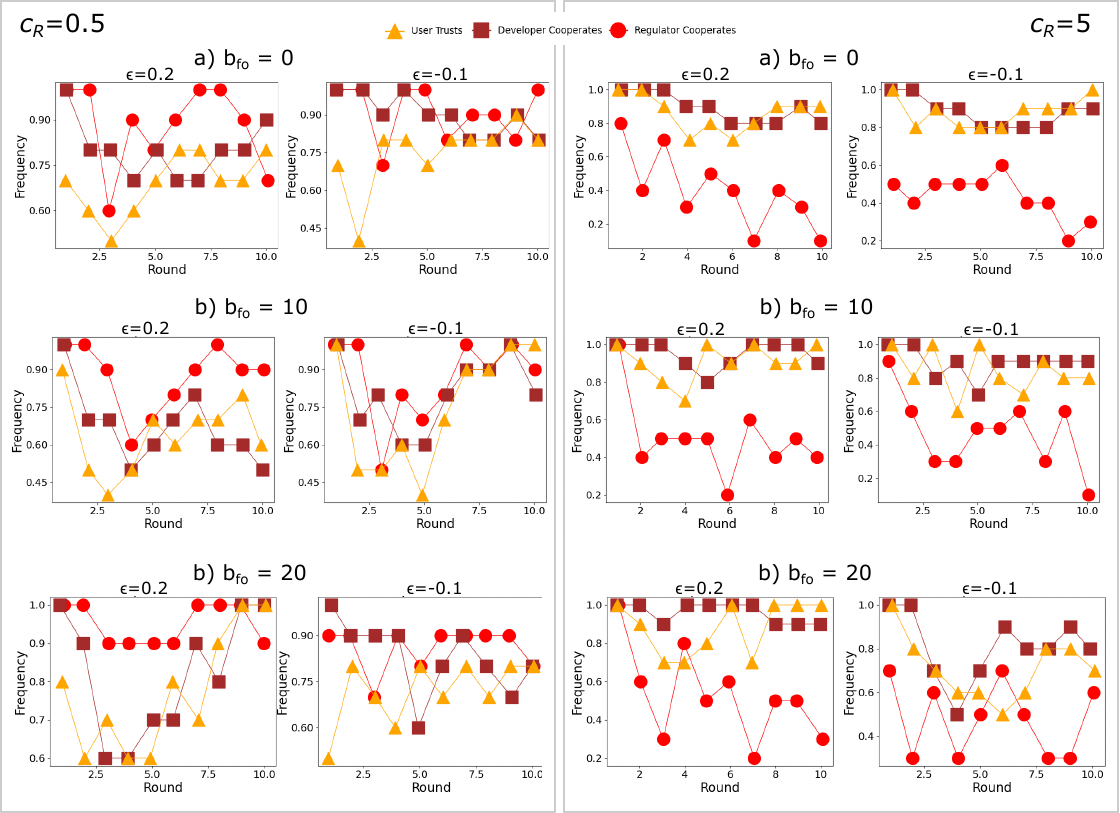}
\caption{Results for the repeated games over each of the 10 rounds, for different values of $b_{fo}$, using Mistral Large Left box: low regulation cost ($c_R$ = 0.5). Right box: high regulation cost ($c_R = 5$). Each panel corresponds to a different value for $\epsilon$, \textit{i.e.}, the risk for users to adopt unsafe AI ($\epsilon<0$ has higher risk). }
\label{fig:repeated_Mistral_rounds}
\end{figure}

We now consider the possibility of having repeated games, such that the proportion of strategies may evolve over several rounds. This way, agents can update their choices based on other players' behaviours in previous rounds, thus becoming able to conditional decisions even in the absence of CT (which is a first approximation of history-dependent choices, and whose results are shown in Supplementary Figure S1). Except for the number of rounds, all other parameters are set as above. Fig. \ref{fig:repeated_both} shows the results as the average over 10 repeated rounds, respectively for GPT- and Mistral-based agents. The results over each round, for selected values of $b_{fo}$, are reported in Fig. \ref{fig:repeated_GPT_rounds} and Fig. \ref{fig:repeated_Mistral_rounds} for GPT-4o and Mistral, respectively; results for each round and each $b_{fo}$ are reported in Supplementary Figs. S3, S4, S6 and S7.

We immediately observe that, for GPT-4o, having repeated games changes the outcome. If the game is repeated among the same group of (three) agents, they end up choosing a mix of strategies, with a tendency to trust (\textit{TCC}) for low regulatory costs ($c_R$), low risk for users $\epsilon$ and low incentives $b_{fo}$, with higher probability of not trusting if $b_{fo}$ increases. Instead, if the regulatory cost is high, there is a higher tendency of not trusting, having defective developers and a mix of complying and defective regulators, whose fraction of compliance increases with higher $b_{fo}$. Overall, the picture that emerges is more in line with what is predicted by game theory using one-shot games (see Figs. 6 and 7 in \cite{alalawi2024trust}): apparently, despite fluctuations given by the stochastic nature of LLMs, repeating the games allows too "smooth out" the effect of data and to converge towards results primarily driven by the payoff matrix.

In Fig. 5 we show, for GPT-4o,  the frequencies of users' trust and developers' and regulators' cooperation over the round, to examine how these players change their behaviour over time. 
We observe that across all scenarios, both developers and regulators start with high levels of cooperation, which tend to decrease  over time. Instead, users tend to trust less, and maintain similar levels of (low) trust over the rounds. The same patterns can be observed with similar trends over all $b_{fo}$ and $\epsilon$, while the absolute values change slightly when changing such parameters.


On the other hand, Mistral Large agents maintain an "optimistic" attitude, preferring to trust as users and to comply as developers and regulators, with high $b_{fo}$ further incentivising regulators to comply in case their cost is negligible, and to be more lenient if the cost of regulating is high and the other actors are already well-behaving. Even in the case of conditional trust, Mistral Large is less sensitive to having repeated games than GPT-4o is. In fact, Mistral's results are in line with the one-shot scenario without CT, where an TCC strategy prevails. Like for GPT-4o, we also observe the emergence of an alternative strategy, namely (C)TDD, where users tend to place their trust upon AI even if developers and regulators are defective. These results, which are close to the game-theoretical predictions in case of \textit{CT} and high $c_R$, are also repeated for low $c_R$, somehow suggesting lower sensitivity of the LLM to this parameter.

In Fig. 6 we show, for Mistral Large,  the frequencies of users' trust and developers' and regulators' cooperation over the round. 
We observe that across all scenarios, the levels of user trust and both developers and regulators tend to decrease  at the beginning and then increase after a few rounds. This is rather in contrast to results obtained with GPT, suggesting that Mistral Large may contain different biases obtained from the training procedure, that tend to prefer cooperative behaviours. These patterns are also in stark contrast to Mistral Large outputs in one-shot games with CT, where a NDD strategy prevails; this suggests that CT, for this LLM, is a poor approximation of updated behaviours due to observation of other agents.

A breakdown on the preferred strategies for each type of agent, average over the various rounds, is also provided in Supplementary Figs. S1 and S4.

\subsection{Adding personality traits}
\label{sub:personality_results}

Finally, we exploit the greater flexibility provided by LLM agents, compared to game-theoretic entities, to isolate the effect of the personality associated to each agent. As described in Sec. \ref{sec:AI_ag_setup}, we prompt the LLMs to play each agent, in one-shot games, according to a set of realistic personalities. We do it for one agent at a time, leaving the other two with personality *None* and then changing the combination; this way, we carefully analyse the role of each agent's personality. This additional set of tests allows us to better interpret the above results by explicitly considering the impact of personalities in the emergent behaviours (recall that, previously, we used the default "personality" that an LLM statistically associates with each agent, that is unknown) and to predict the impact that specifying a personality has on the strategy choices by AI agents. For each agent, we use the set of personalities described in Sec. \ref{sec:AI_ag_setup}.

Due to resource constraints, we focus on the one-shot game scenario with conditional trust, $c_R=0.5$ and $\epsilon=-0.1$, which  showcases the most interesting outcomes according to \cite{alalawi2019pathways} and to the findings above. All other parameters are set as previously.

\begin{figure}[h]
\includegraphics[width=\linewidth]{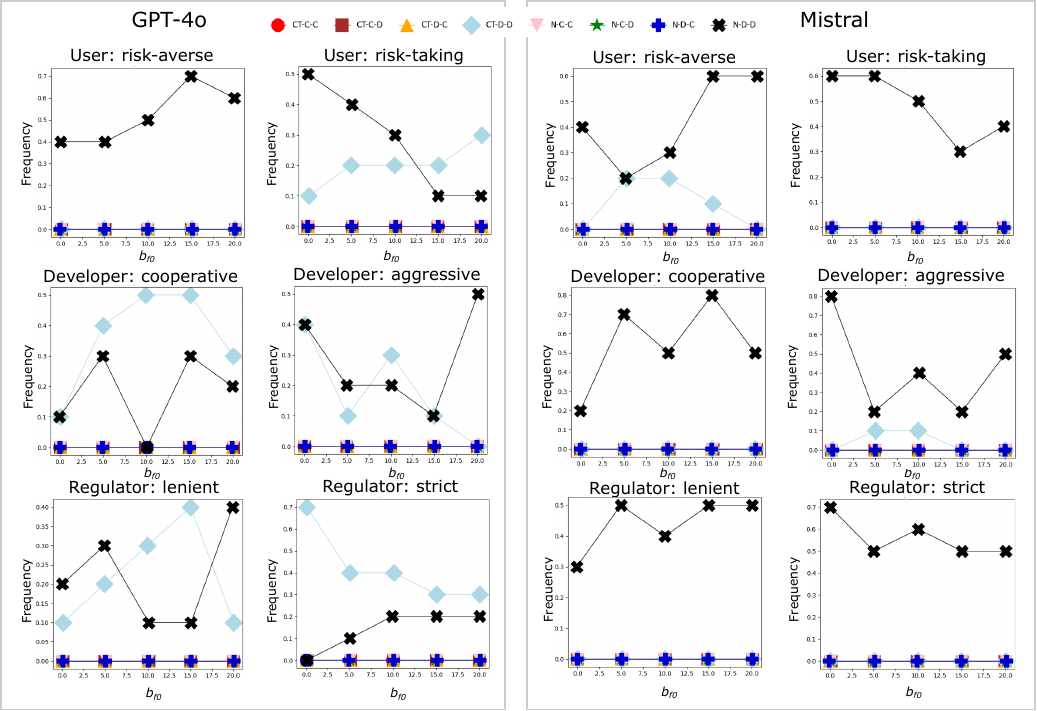}
\caption{Results for the one-shot games with personality traits, using GPT-4o (left) and Mistral Large (right). Each row contains a set of two contrasting personalities, described in \ref{sec:methods}. All scenarios are with \textit{CT}, $c_R=0.5$ and $\epsilon=-0.1$.}
\label{fig:oneshot_personality}
\end{figure}

Fig. \ref{fig:oneshot_personality} summarises the results for both GPT-4o and Mistral Large. Additional results are in Supplementary Fig. S7. From Fig. \ref{fig:oneshot_personality}, we immediately see that the personality results align with the repeated games and Mistral one-shot outcomes. Except for when developers cooperate and users are more risk-taking, where conditional trust by users may emerge despite defection by the other players, all the other scenarios maintain NDD as the main strategy. An alternative to promote trust in users is picked by GPT-4o when regulators have their own personality, which may affect users' trust on top of reputation. In general, GPT-agents are more optimistic than Mistral's ones, which mostly replicate the "pessimistic" results obtained without personalities, \textit{cf.} Fig. \ref{fig:one_shot_Mistral}.

Overall, equipping AI agents with personalities offer an additional array of nuances and possible outcomes, that can inform predictions about trust in AI regulation. When personalities are specified, both LLMs provide similar results, suggesting that fixing this extra parameter is key to increase the chance of repeatable outcomes through LLMs. Moreover, the preset results suggest that AI agents are, overall, aligned towards a rather pessimistic view of current AI regulation, if sustained by conditional trust. This observation may suggest that, when personality is set to *None*, the default personality of AI agents is more closely aligned towards "pessimistic" attitudes. Understanding whether this is systematic, and whether it emerges from training data, is demanded to future studies.

\section{Discussion}

This work provides a first systematic test of AI agents within a game-theoretic framework, for a game that is not standard in the literature and thus have no data to train the LLMs upon. The obtained results are thus genuinely deriving from the interaction of payoff matrices and \textit{any other type of data} used for training, while previous results \cite{lu2024llms,buscemi2025fairgame} potentially contained pre-cooked outputs deriving from games included in the training of LLMs. Our results thus provide crucial insights according to two perspectives. On the one hand, what can we learn about strategic decisions on AI regulation, from the perspective of AI agents (and predicting how they would behave if they are given access to AI governance decisions as suggested for other industries \cite{chen2024intelligent}). On the other hand, they allow to better interpret LLM responses under the lenses of a well-established theory, thus advancing our capabilities to advance hypotheses about their inner functioning and promoting the development of complex-systems models for AI interpretability. \\

About the question \textit{Do LLM trust AI regulation?}, our study provides several key takeaways. 

Overall, AI-users place their trust upon developers and regulators depending on their behaviour and, potentially, on their personality (despite the latter seems to play a minor role). Then, trust depends radically on the used LLM: for GPT-based agents, conditional trust promotes overall trust, while Mistral agents experience the exact reverse situation in case of one-shot games. Repeated games make the LLMs more aligned, in that they predict that users would have mixed or relatively high trust if not conditioned by regulators' reputation, but would not trust in case of \textit{CT}. Recalling that LLMs likely mix payoff-based games with statistical outcomes deriving from their training upon real data, this may suggest that regulators have relatively low reputations on the data sources, and thus LLM agents tend to trust them less. If the risk of unsafe AI development is significant, regulatory authorities should thus show high dependability to promote trust in users. 

When the cost of regulation is low, and regulators tend to regulate more, AI agents suggest that developers would tend to comply more, but may defect otherwise. Instead, regulators are suggested to have the tendency of defecting, unless properly incentivised. Regulatory bodies should thus guarantee manageable regulatory costs and have a high capacity to identify non-compliant, in order to consistently enforce safety measures.

In general, however, the results are less clear-cut than those obtained by pure game theory, but present more nuances, potentially associated with the intrinsic randomness of LLMs, as well as with the presence of biases in the data that may balance payoff-based decisions. These results complement previous calls to urgently develop actions towards AI safety and trustworthiness, and to allocate the necessary resources to monitoring bodies \cite{kinniment2024evaluating, k2023institute}. Also, they suggest the need to include ethical and safety considerations in the governance discourse to enable trustworthy and human-centric AI that can promote trust among users, and warn against attempts to automatise the governance processes, especially for AI development and regulation. \\

Then, we observe that LLMs are not completely aligned with theoretical game results. This can derive from interferences from training data, from the challenges that LLMs encounter when performing mathematical modelling while primarily working with statistical associations \cite{ahn2024large}, or by their different sensitivity to elements of the payoff matrix. Nonetheless, this observation opens exciting avenues: identifying the best LLMs to embed elements of empirical data may elicit game studies that also reflect real-world situations and possibly improve predictions. However, this endeavour may be tackled very carefully, to avoid spurious results deriving from the black box nature of LLMs. In fact, the panels in Fig. \ref{fig:oneshot_personality} follow closely the corresponding panels of Fig. \ref{fig:repeated_both}; instead, only the one-shot scenario for GPT-4o (Fig. \ref{fig:one_shot_GPT}) yields conditional trust as predicted by game theory. This observation can be interpreted under two hypothesis, whose verification will support the development of future studies: either LLMs are more biased by training data, and have the tendency to capture polarization and mistrust in institutions and developers, or they effectively embed more nuances that game-theoretic models, which would thus constitute best-case scenarios. In both cases, merging game theory and LLM is suggested as a powerful  avenue to improve predictions about strategic behaviours.

Finally, this study uncovers a methodological caveat for game theory scholars who approach LLM-based simulations, that is, the selection of one LLM or another has a profound impact in the results, similarly to having different samples of populations to perform human experiments \cite{hagen2006game}. To ensure reproducibility and reliable predictions, additional tests on the behaviour of LLMs, and improving their interpretability and selection guidelines, are strongly recommended to blend the best of both disciplines and unlock new research avenues with profound social impact and immediate applications for the development of strategic agents in social, governance and economical ecosystems.

\subsection{Limitations and avenues for future research}

Our model embeds essential mechanisms for strategic decision-making and, thanks to FAIRGAME, ensures the reproducibility of the results. However,  several limitations remain, that can promote future research and more refined insights. First, we considered mean-field behaviours for users, developers and regulators --  and even mean-field personalities. In reality, users are segmented into market niches that choose which AI system they want to use, and companies may choose to relocate to avoid especially burdensome regulation. On the other hand, compliance with AI safety regulations may become a value proposition for companies, whose cost would be drastically reduced thanks to the competitive advantage gained. Future research could address these limitations by incorporating networks or heterogeneous populations, or partner selection between populations, and potentially considering non-linear cost structures. Moreover, except for the conditional trust argument, we do not include feedback loops between the perception of all agents; state-dependent payoff weights may address this point. 

Another area for future developments is to model more explicitly the competition between different regulatory agencies, often racing for resources and group selection \cite{richerson2016cultural, van2009group}. Network-based or agent-based approaches may shed light onto their effects. Similarly, different developers are in fierce competition with one another and, as the market currently stands, pursuing the frontier of AI prowess has higher priority than pursuing AI safety research agendas \cite{armstrong2016racing, cottier2024who}. However, new market segments are opening, also under the pressure of regulatory bodies, and diversification in the AI products may also include higher attention for safety and transparency (at least, as narratives) as in the case of the startup Anthropic. As suggested earlier, including market competition may enrich the insights. 

Moreover, we have simplified the flow of information and regulation across agents. Usually, media and academia assume the role of conveying and interpreting information about sentiment and trust across the actors involved. Similarly, the developers' job is usually filtered to users through marketing or other vendors, thereby shadowing the hidden content of many AI systems. Regulatory bodies also often rely on agencies and accountants to survey and report. Future studies may explicitly consider these mediating effects by adding a population of multiple LLM agents, assuming each role. 

We also comment on the LLMs themselves. Testing the capabilities of AI agents to interpret leading roles in strategic games is crucial to prepare for their use in different applications, and may shed light onto non-linearities and complexities that are necessarily overlooked by simpler modelling approaches. However, the black-box nature of LLMs require careful evaluation and interpretation of the results. As studies to improve their interpretability proceed \cite{ali2025entropy, el2025towards}, future research will dig deeper into the determinants of emerging behaviours by AI agents playing games. In this work, we have provided interpretations based on knowledge of LLM training and functioning, but they should be considered hypothesis to be tested with additional tools. Future research should test the predictions of our model and prove evidence of strategic interactions between users, developers, and regulators in the AI domain. As discussed earlier, the cross-talk between LLM research and formal disciplines such as game theory have the potential to uncover hidden biases and stimulate hypotheses. Developing a new LLM-oriented game-theoretic framework, embedding statistics and optimization problems to explain and predict emerging strategic behaviours by AI agents, may further propel this area of research. \\

In general, despite its modelling limitations, our simple scenario is useful to think about which assumptions should policymakers make, to promote adoption of safely developed  AI. Crucially, we have investigated what game-based AI systems would suggest about such assumptions, challenging their trust on the behaviours and strategic decisions of the main actors of the AI landscape. This inception may have value in stimulating reasoning and predictions, as well as to inform strategic decision-making based on complex models.

\section*{Acknowledgement}
This work was produced during the workshop "AI Governance Modelling", funded through the generous support from the Future of Life institute (T.A.H).
T.A.H. and Z.S. are supported by EPSRC (grant EP/Y00857X/1). M.H.D  and N.B. are supported by EPSRC (grant EP/Y008561/1) and a Royal International Exchange Grant IES-R3-223047.
E.F.D. is supported by an F.W.O. Senior Postdoctoral Grant (12A7825N),
A.M.F. and H.C.F. were supported by INESC-ID and the project CRAI C645008882-00000055/510852254 (IAPMEI/PRR). D.P is supported by the European Union through the ERC INSPIRE grant (project number 101076926); views and opinions expressed are however those of the authors only and do not necessarily reflect those of the European Union, the European Research Council Executive Agency or the European Council.

\bibliographystyle{IEEEtran}
\bibliography{refs} 

\end{document}